\documentclass[]{article}
\usepackage[letterpaper]{geometry}
\usepackage{mtsummit2017}
\usepackage{times}
\usepackage{url}
\usepackage{latexsym}
\usepackage{layout}
\usepackage{verbatim}
\usepackage{multirow}
\usepackage{graphicx}
\usepackage{booktabs}
\usepackage{subcaption}
\usepackage{footmisc}
\usepackage{amssymb}
\usepackage[numbers]{natbib}

\begin{document}
\title{\bf Toward a full-scale neural machine translation in production: the Booking.com use case}
\author{
	\name{\bf Pavel Levin} \hfill  \addr{pavel.levin@booking.com}\\ 
        \name{\bf Nishikant Dhanuka} \hfill \addr{nishikant.dhanuka@booking.com}\\ 
        \name{\bf Talaat Khalil} \hfill \addr{talaat.khalil@booking.com}\\                 
        \name{\bf Fedor Kovalev} \hfill \addr{fedor.kovalev@booking.com}\\         
        \name{\bf Maxim Khalilov} \hfill \addr{maxim.khalilov@booking.com}\\
}

\maketitle
\pagestyle{empty}

\vspace{0.15in}
\begin{abstract}
While some remarkable progress has been made in neural machine translation (NMT) research, there have not been many reports on its development and evaluation in practice. This paper tries to fill this gap by presenting some of our findings from building an in-house travel domain NMT system in a large scale E-commerce setting. The three major topics that we cover are optimization and training (including different optimization strategies and corpus sizes), handling real-world content and evaluating results.

\vspace{0.1in}
 

\end{abstract}

\section{Introduction}
Booking.com is one of the largest online companies in the world operating in 43 different languages, connecting millions of daily visitors to 1.4 million bookable accommodations while offering both parties multilingual support and information every step of the way. Given the company's fast growth and a rising need for more high quality translated content, machine translation (MT) is becoming an increasingly attractive option to automate this difficult task.

Our experiments \cite{levin2017a} consistently show the superiority of neural machine translation (NMT) systems over the more traditional statistical ones, even when we benchmark them against the well-established and tested general purpose systems. Therefore our recent focus has been on tailoring and improving our own in-house NMT systems to make them practical and effective for us. This work highlights some of the main learnings on our journey and should be of interest to anyone looking to deploy a custom NMT system.

In particular we focus on the following three major topics:

\begin{itemize}
\item \textbf{Optimization and training}

At Booking.com we have collected tens of millions of travel domain specific human-translated parallel sentences, which in theory allows us to train very flexible models with hundreds of millions of parameters. However learning such system can be computationally expensive which often translates to unacceptably long product development iteration cycles. To address this we first analyze how convergence is affected by different optimization techniques (Section \ref{sect:opt_fit}), including in a multi-GPU environment. Second, we look at how the quality of a trained system improves as a function of the training corpus size (Section \ref{sect:opt_learn}).
\item \textbf{Handling real-world content}

Real world text comes with many challenges which have to be addressed. Section \ref{sect:rw} presents some practical considerations for dealing with named entities and rare words.
\item \textbf{Quality evaluations}

When building an MT system with customer-facing output, setting up a good quality evaluation loop can be one of the most important aspects. In this part we show how in addition to the BLEU metric \cite{Papineni02bleu:a}, the de facto standard for automatic MT scoring, we employ human evaluation of translation adequacy and fluency. We take a close look at how the two approaches correlate. Further, we share our experience developing our business sensitivity framework, which helps us proofread the final translation identifying particularly pernicious errors.
\end{itemize}

\section{Optimization and training}\label{sect:opt}

\subsection{Model architecture}
The core of our translation pipeline is based on OpenNMT \cite{2017opennmt}, which is a Lua written framework for training encoder-decoder neural architectures. Usually, both the encoder and the decoder recurrent neural networks (RNNs), in our case typically long short-term memory (LSTM) units \cite{lstm}, each with 4 layers. We always use (global) attention layer with input feeding to help the model learn faster by keeping a ``memory" of past alignment decisions \cite{luongeffective}. For European languages we use ``case features" (see Section \ref{sect:tok}) as additional input variables from the ``cases" embedding space \cite{sennrich2016linguistic}. The main word embeddings are concatenated with the case embeddings to form the inputs to the encoder. At each layer of the encoder the RNNs are bi-directional \cite{schuster1997bidirectional}. Both the encoder and the decoder use residual connections between layers \cite{he2016deep} as well as the dropout rate of 0.3 \cite{srivastava2014dropout}.

\subsection{Optimization and model fitting}\label{sect:opt_fit}
\subsubsection{Single-GPU environment}
\begin{figure}[h]
\centering
\includegraphics[scale=0.275]{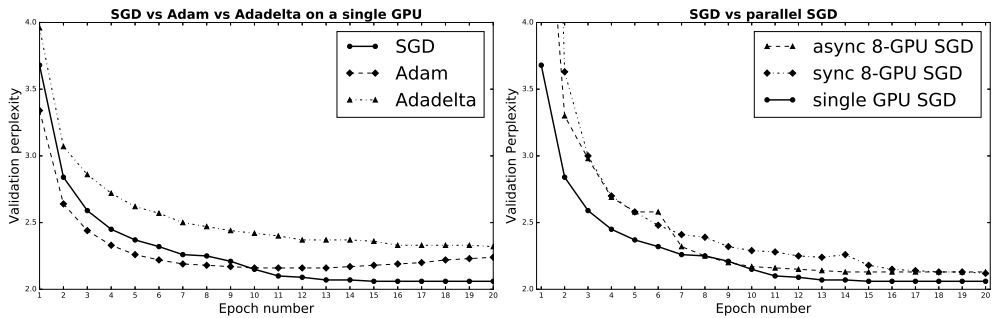}
\caption{Model convergence. The subplot on the left shows model convergence for three different optimizers: SGD, Adam and Adadelta. Adagrad in our setting did so poorly that it would not fit in the plot (its validation at epoch 20 was above 12). The right subplot compares the convergence of SGD on a single GPU to those of SGD run on an 8-GPU cluster using synchronous and asynchronous parameter updates.}\label{fig:opt_exp}
\end{figure}

To optimize the training of our NMT system in single-GPU environments, we evaluated different algorithms primarily based on their speed of convergence and translation output quality. The dataset used was English-German property descriptions with one million parallel sentences. We conducted experiments with four well-known optimizers: stochastic gradient descent (SGD) with learning rate decay, Adam \cite{kingma2014adam}, Adagrad \cite{duchi2011adaptive} and Adadelta \cite{zeiler2012adadelta}. Our SGD decay strategy is based on a combination of the perplexity score and epoch number, meaning we decay current learning rate by a multiplicative factor of 0.7 if current epoch's validation perplexity does not decrease, and after each epoch after the 9th epoch. Our initial learning parameters for SGD, Adam, Adagrad and Adadelta are 1.0, 0.0002, 0.1, and 1.0 respectively. We ran the model for 20 epochs and used both perplexity per epoch and BLEU score after every five epochs on the validation set of 10,000 sentences to measure the performance. Our results are summarized in Table \ref{tab:opt_exp} and Figure \ref{fig:opt_exp}.

\begin{table}[h]
\centering
\begin{tabular}{|@{\hskip3pt}c@{\hskip3pt}||c|c|c|c@{\hskip4pt}||c|c|c|c@{\hskip2.5pt}||c@{\hskip3pt}|}
 \multirow{2}{*}{\textbf{Optimizer}}& \multicolumn{4}{|c||}{\textbf{Perplexity}}  & \multicolumn{4}{|c||}{\textbf{BLEU}} &\multirow{2}{*}{\begin{tabular}{@{}c@{}}\textbf{Time} \\ \textbf{per epoch}\end{tabular}   }  \\ \cline{2-5} \cline{6-9}
& \textbf{5} & \textbf{10} & \textbf{15} & \textbf{20} & \textbf{5} & \textbf{10} & \textbf{15} & \textbf{20} &  \\ 
\hline
\begin{tabular}{@{}c@{}}SGD \\ with decay\end{tabular} & 2.37 & \textbf{2.15} & \textbf{2.06} & \textbf{2.06} & 43.74 & 45.10 & \textbf{45.84} & \textbf{46.58} & 6h 11m  \\
Adam & \textbf{2.26} & 2.16 & 2.18 & 2.24 & \textbf{44.89} & \textbf{45.33} & 45.21 & 44.78 & +40m \\
Adagrad & 38.75 & 19.82 & 15.21& 12.55 & 1.4 & 2.25 & 2.56 & 3.14 & +14m\\
Adadelta & 2.62 & 2.42 & 2.36 & 2.32 & 42.43 & 43.42 & 44.35 & 44.07 & +54m
\end{tabular}
\caption{Performance of different optimizers on training English-German translation model reported every 5 epochs. Each experiment was conducted in a single NVIDIA Tesla K80 GPU.}\label{tab:opt_exp}
\end{table}

As can be seen in Table \ref{tab:opt_exp}  and Figure \ref{fig:opt_exp}, we observed that initially Adam converged faster as expected because it applies momentum on a per parameter basis, but SGD took over as soon as decay started and outperformed Adam thereafter. The perplexity reached by SGD in the 9th epoch was already achieved by Adam in the 6th. But from the 10th epoch onward, as soon as SGD learning rate starts decaying indefinitely, Adam's perplexity is consistently worse than that of SGD. However, there was no decrease in perplexity from 15th till 20th epoch, so SGD already converged by epoch 15. We also observed that Adagrad performed very poorly on our model. Adadelta was much better than Adagrad but still slightly behind Adam and SGD. We further validated our results using BLEU scores every 5 epochs. The results were mostly consistent with what we observe by looking at perplexity. In terms of time taken per epoch, SGD was the fastest. Adam was about 10\% slower in comparison.

\subsubsection{Multi-GPU environment}
Next we experimented with the use of multiple GPUs by using data parallelism technique which trains batches in parallel on different GPUs. On a single GPU our model takes 6h11m per epoch on average, and we usually see it converging around 15th epoch, which means training a model on only 1 million sentences takes about 4 days. 15 epochs on a corpus of size 10M could easily translate to around 40 days\footnote{Reported estimates do not account for any time related to model checkpointing.}. In an attempt to speed up our development cycle, we ran some experiments with synchronous and asynchronous SGD (with decay) on a cluster of 2, 4, 6 and 8 GPUs. The main difference between these two approaches is that in synchronous mode all gradients are accumulated and parameter updates are synchronized, while in asynchronous each GPU calculates its own gradient and communicates with the ``master copy" of parameters independently and asynchronously. This mater copy of parameters is stored on a single dedicated GPU which is not used for training. To achieve a faster convergence through better parameter initialization, only one GPU works for the first 6,000 iterations in async SGD.

\begin{figure}
\begin{tabular}{cl}
\centering
  \raisebox{-.5\height}{\includegraphics[scale=0.35]{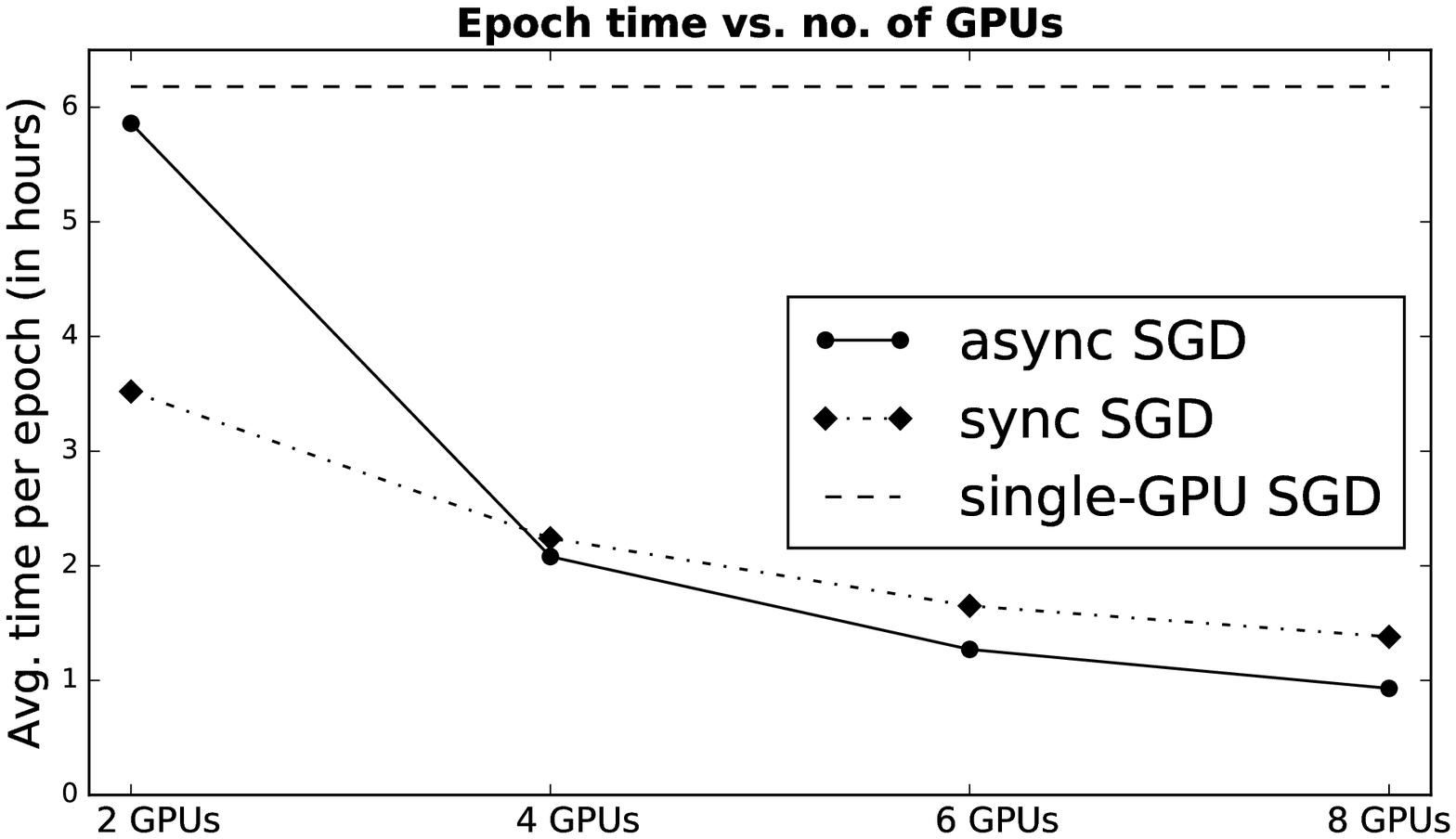}} &
  \hspace{-0.5in}
\begin{tabular}{r||c|c|r}
{\small \textbf{GPUs}} & \textbf{sync SGD} & \textbf{async SGD}  \\
2 & 3h31m & 5h52m \\
4 & 2h14m & 2h05m \\
6 & 1h39m & 1h16m \\
8 & 1h23m & 56m \\
\end{tabular}
\end{tabular}
\caption{Time per 1M iterations taken by synchronous and asynchronous SGD. On a single GPU the same model takes 6.18 hours.}\label{fig:opt_multi}
\end{figure}

As can be seen in Figure \ref{fig:opt_multi}, average time per epoch came down as we added more hardware: from 6h11m to 1h23m for sync and to only 56 minutes for async. Note that with 2 GPUs, async takes almost the same time as non-parallel SGD (around 6 hours) while sync is much faster at 3h31m. The reason for that is that 2-GPU async is almost equivalent to a single GPU model as async blocks one GPU completely to store the master copy of parameters and is not used for training. Because async mode skips the overhead of parameter synchronization, it was expected that it would be faster than sync, so we also looked at the quality as measured by perplexity. 
During the first epoch sync perplexity is much worse than that of async due to only 1 GPU working in async for first 6,000 iterations resulting in better parameter initialization (this cannot be seen in Figure \ref{fig:opt_exp} which has been cropped for better visibility; sync has first epoch perplexity of 9.61, compared to 5.61 for async and 3.68 for single-GPU SGD). However, for all remaining epochs their scores are very similar. Single-GPU SGD, on the other hand, performed noticeably better in the first half of the training, but gets quite similar to multi-GPU models eventually (although still marginally better). Overall we are very happy with async's performance as it is able to reduce the training time by about 85\%.

\subsection{The importance of corpus size}\label{sect:opt_learn}

\begin{figure}
\begin{tabular}{cl}
\centering
  \raisebox{-.5\height}{\includegraphics[scale=0.45]{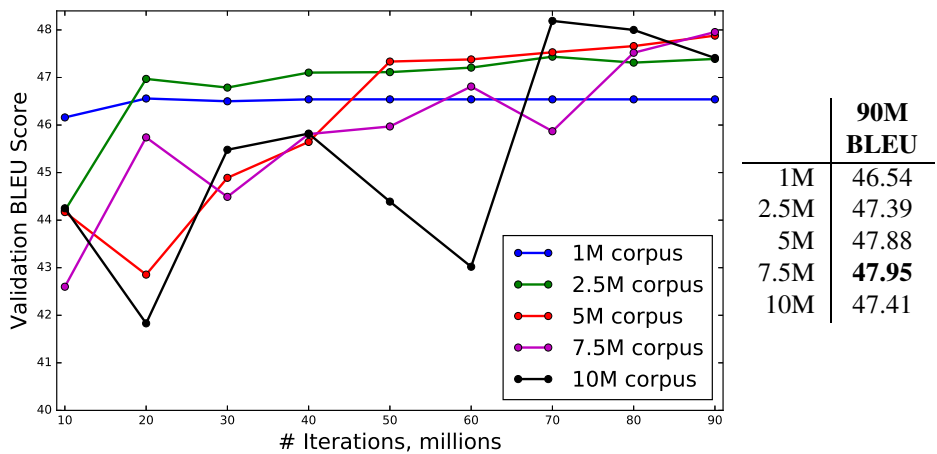}} &
  \hspace{-0.55in}
\begin{tabular}{r|c}
 & \begin{tabular}{@{}c@{}}\textbf{90M}\\ \textbf{BLEU}\end{tabular} \\
 \hline
1M & 46.54  \\
2.5M & 47.39  \\
5M & 47.88  \\
7.5M & \textbf{47.95} \\
10M & 47.41 \\
\end{tabular}
\end{tabular}
\caption{Performance (measured by BLEU score) of a model trained on different corpus sizes, reported every 10M iterations.}\label{fig:train_curve}
\end{figure}

In order see how much benefit we get from an increased corpus size, we compared models trained on 1M, 2.5M, 5M, 7.5M and 10M sentences. For fair comparison we report the learning curves as a function of number of iterations (training time) and not the epoch number. Figure \ref{fig:train_curve} shows our findings.

Essentially there were no major surprises. It appears that given enough iteration the model with more distinct sentences will have a higher BLEU score. Notice how in the beginning smaller datasets are actually winning, but given enough training time the model is starting to take full advantage of more data. The largest corpus size of 10M does not have the best performance at the end of 90M iterations, however as we shall see in Section \ref{sect:q_bleu} this is in fact not true and according to human evaluations 10M gives the best results which are simply not captured by the BLEU metric.

\section{Handling real-world content}\label{sect:rw}

\subsection{Tokenization and case features}\label{sect:tok}

In our final models we use byte-pair encoding (BPE) tokenization procedure \cite{sennrich2015neural}. BPE is a compression technique which was recently adapted to find optimal tokens for sequence composition in sequence-to-sequence learning tasks. In theory the technique should find a perfect compromise between using word-level translation (and dealing with out-of-vocabulary entities) and character-level translation (and dealing with much longer sequences of tokens). The procedure is very straightforward. We start with a set of tokens which is the list of acceptable characters and iteratively grow it, at each step adding a concatenation of two items already in the list which is the most frequent in our corpus. The number of iterations can be viewed as the algorithm's only hyperparameter. We can either apply BPE to the source and the target sentences separately, or we can apply them to the combined corpus. Based on our experiment (see Table \ref{tab:bpe_exp}) we decided ended up with the joint version.

\begin{table}[h]
\centering
\begin{tabular}{r||c| c|c|c|c|| c| c| c| c}
&\multirow{2}{*}{\begin{tabular}{@{}c@{}}\textbf{50k-Vocab} \\ \textbf{baseline}\textsuperscript{\cite{levin2017a}}\end{tabular}   } & \multicolumn{4}{c||}{\textbf{Joint BPE}}  & \multicolumn{4}{|c}{\textbf{Separate BPE}}\\
&  & 30k & 50k & 70k & 90k & 30k & 50k & 70k & 90k \\
\hline
Epoch\ \ \ 5 &  39.54 & 43.75 & 43.46 & 43.40             & 41.23 & 42.81 & 42.35 & 39.73& \multirow{4}{*}{{\textbf{N/A}}}\\
Epoch 10 &     40.95 & 44.55 & 44.52 &43.81             & 43.81 & 43.39 & 43.48 & 43.51 & \\
Epoch 15&      42.01 & 45.08 & 45.91 &46.14             & 45.75 & 43.58 & 43.23 & 45.17 & \\
Epoch 20 &     42.15 & 46.31 & 46.43 &\textbf{46.61} & 45.62 & 45.22 & 46.00 & 45.90 & \\

\end{tabular}

\caption{Comparison of the BLEU scores of identically trained models with different BPE configurations, as well as the baseline with a vocabulary of 50,000 most common words (see \cite{levin2017a} for more details on the baseline model). All experiments were run on 1M corpus. We found 70,000 tokens (70k) jointly trained BPE to have the highest validation BLEU score. Because we saw a strong pattern which made it clear that separately trained BPE 90k model was not going to win, we decided to not run that experiment as it is also the most expensive one.}\label{tab:bpe_exp}
\end{table}

Apart from applying BPE tokenization we also use case features preprocessing. This allows us to map the same words and word pieces spelled with different cases to the same embeddings while also passing the casing information separately. For example raw terms \textit{book}, \textit{Book} and \textit{BOOK} would all be mapped to the same token \textit{book}, but would have different accompanying case feature values. Case features get their own embeddings which get combined with token embeddings during the translation \cite{sennrich2016linguistic}. In theory this greatly increases the encoding and decoding efficiency of the system, which we also observed in practice through much better performance over not using case features.

\begin{table}[h]
\centering
\normalsize

\bgroup
\def\arraystretch{1.3}
\begin{tabular}{l||l}
\hspace{-0.12in}
{Raw source}
&\hspace{-0.05in}{  \begin{tabular}{@{}l@{}}{\large Offering a restaurant with WiFi, Hodor Ecolodge}\\ {\large  is located in Winterfell.}\end{tabular}}\\
\hline
\hspace{-0.12in}
{Tokenized source}
 &\hspace{-0.05in}{ \begin{tabular}{@{}l@{}}{\large offering\textsuperscript{C} a\textsuperscript{L} restaurant\textsuperscript{L} with\textsuperscript{L} wi{\small $\blacksquare$}\textsuperscript{C} fi\textsuperscript{C} {\small $\blacksquare$},\textsuperscript{N} ho{\small $\blacksquare$}\textsuperscript{C}}\\ {\large dor\textsuperscript{L} ecolodge\textsuperscript{L} is\textsuperscript{L} located\textsuperscript{L} in\textsuperscript{L} winter{\small $\blacksquare$}\textsuperscript{C} fell\textsuperscript{L} {\small $\blacksquare$}.\textsuperscript{L}}\end{tabular}}\\
\hline
\hspace{-0.12in}
{Tokenized Output} &
\hspace{-0.05in}
\begin{tabular}{@{}l@{}}
{\large die\textsuperscript{C} ho{\small $\blacksquare$}\textsuperscript{C} dor\textsuperscript{L} ecolodge\textsuperscript{C} in\textsuperscript{L} winter{\small $\blacksquare$}\textsuperscript{C} fell\textsuperscript{L} bietet\textsuperscript{L}}\\
{\large ein\textsuperscript{L} restaurant\textsuperscript{L} mit\textsuperscript{L} wlan\textsuperscript{U} {\small $\blacksquare$}.\textsuperscript{N}}
\end{tabular}\\
\hline
\hspace{-0.12in}
{De-tokenized output}
&
\hspace{-0.05in}
\begin{tabular}{@{}l@{}}
{\large Die Hodor Ecolodge in Winterfell bietet ein}\\
{\large Restaurant mit WLAN.}
\end{tabular}
\end{tabular}
\egroup
\caption{A typical sentence describing an accommodation translated from English to German. Before being fed into the encoder, the sentence is first tokenized using byte-pair encodings. Notice how the words ``Hodor" and ``Winterfell" which never occurred in our training corpus are split into pieces which are understood by the encoder. The symbol {\small $\blacksquare$} indicates no space between two neighboring word pieces. The superscripts are case features (C: true case, L: lower case, U: all capitals, N: non-alphabetic)}
\end{table}

\subsection{Handling named entities}\label{sect:ner}
Text in the travel domain contains a large amount of entities. There is almost always some destination involved, a property name, distances, times, etc. Although many NMT researchers report results on end-to-end neural networks \cite{Cho2014,crego2016systran,Wang2017}, we often found RNN encoder-decoder architecture insufficient to produce acceptable results, mainly due to mishandled named entities. This section outlines our approach to processing such entities which drastically improves the translation output quality.

As an example, mistranslated distances constitute one of the most common error types when NMT is applied naively on raw text, even with very large corpus sizes (over 10M parallel sentences). Interestingly NMT often correctly converts between kilometers and miles for commonly occurring distances (e.g.\ 5km, 10 miles); however, the number of distance-related mistakes in our validation set is too large to be left untreated. Another common type of error is related to times and dates (12 vs 24 hour clock times, different date formats).

\begin{table}[h]
\centering
\normalsize

\bgroup
\def\arraystretch{1.3}
\begin{tabular}{l||l}
\hspace{-0.12in}
{Source sentence}
&\hspace{-0.05in}{  \begin{tabular}{@{}l@{}}{\large Winterfell Railway Station can be reached in a \textbf{55-minute}}\\{\large \textbf{car ride}.}\end{tabular}}\\

\hline

\hspace{-0.12in}
{Pure NMT translation}
 &\hspace{-0.05in}{ \begin{tabular}{@{}l@{}}
{\large Den Bahnhof Winterfell erreichen Sie nach einer}\\
{\large \textbf{5-minütigen Autofahrt}.}
\end{tabular}}\\
\hline

\hspace{-0.12in}
\begin{tabular}{@{}l@{}}
{NMT with distance}\\{placeholders}\end{tabular}
 &
\hspace{-0.05in}
\begin{tabular}{@{}l@{}}
{\large Den Bahnhof Winterfell erreichen Sie nach einer}\\
{\large \textbf{55-minütigen Autofahrt}.}
\end{tabular}
\end{tabular}
\egroup
\caption{Translation of a sentence involving a distance using a BPE-based NMT model and an identically trained model with placeholder preprocessing. These types of errors are critical, however they are not adequately reflected in the BLEU score or decoder perplexity change.}
\end{table}

In most such cases we used a set of manually created templates to search for entities and replace them with special placeholders. As our team does not understand most of the languages that we build MT systems for, we get some help from our in-house language specialists (translators). The template refinement cycle goes as follows. We come up with a set of reasonable regular expressions to identify named entities of a certain type in both languages and run them on our parallel corpus. Then we take the set of sentences where the numbers of recognized entities differs between the source and the target. We then look at the breakdown of most common entities in either language which did not have corresponding parallel counterparts, and refine our regular expressions accordingly. At translation (prediction) stage, we preprocess the input to replace all named entities with corresponding placeholders, run the translation, then substitute back the named entities parsed according to the target language format. This simple approach dramatically improves the translation output quality for sentences which involve problematic named entities.



\section{Quality evaluation}
Unlike simple classification or regression tasks, sequence learning problems are much more difficult to evaluate. The problem comes from the fact that there can be many possible solutions and it is hard (and often impossible) to compare the model output to all valid ``true values". To assess the quality of translations automatically, a useful heuristic is the so-called BLEU score \cite{Papineni02bleu:a} which roughly measures the degree of word overlap between the model translation and a human translation. BLEU score is attractive because it is completely automatic given translated sentences and corresponding model predicted sentences. However, multiple problems have been noted in using BLEU score alone. As a purely counting-based metric, BLEU will favor translation which have more common words and n-grams with the reference translation, regardless of the sentence grammar. It would also penalize models which rephrase the sentence in a way which uses different words from the reference sentence, while preserving its meaning. 

In this section we first describe how we leverage our in-house linguistic expertise to score our models in a relevant way (Section \ref{sect:q_human}). Then we analyze how BLEU score correlates with human metrics (Section \ref{sect:q_bleu}). 

\subsection{Human evaluation loop}\label{sect:q_human}

Our main human evaluation is based on adequacy/fluency methodology\footnote{\url{https://www.taus.net/academy/best-practices/evaluate-best-practices}} which, as the name suggests, is based on two criteria: \textit{adequacy} and \textit{fluency}. Adequacy shows to what degree the meaning of the source sentence is preserved, while fluency scores how grammatically well-formed (from the native speaker's perspective) the translated segment sounds. Each sentence is scored by two independent professional translators from English to German (native German speakers). For the experiments in Section \ref{sect:q_bleu} we chose 200 randomly selected sentences and translators with at least one year of experience professionally translating Booking.com content.

Additionally we use human evaluators to score the quality of entity handling (as described in Section \ref{sect:ner}). For that task each sentence known to contain a specific entity type is given a binary score of whether or not the entity is translated correctly. We found having a separate evaluation specific to entities in addition to adequacy and fluency is important as it helps us to decide on tokenization procedure, entity handling procedures, etc.

\subsection{Business sensitivity analysis}
One important shortcoming of the BLEU score is that it says nothing about the so-called ``business sensitive" errors. For example, the cost of mistranslating ``Parking is available" to mean ``There is free parking" is much greater than a minor grammatical error in the output. Typically it is very difficult to detect such errors because doing so requires some understanding of the sentence \textit{meaning}. Even so, given the potentially huge cost of such mistakes, we have developed a basic ``business sensitivity framework" (BSF) layer to detect certain specific types of errors.

The way it works is rather straightforward. It is a two-stage system, where we first identify the sentences with a particular sensitive aspect (e.g.\ parking availability, pet policy, etc.) then we apply two classifiers (one to the source sentence, the other  to the translation) to identify the predicted values of this aspect (e.g.\ ``free parking", ``pets not allowed" etc.) Finally, BSF flags the sentence as problematic if the predicted aspect values differ between the source and the translation. For the first layer of finding relevant sentences, we learn word and phrase embeddings by training word2vec \cite{NIPS2013_5021} on our full (monolingual) corpora. Then we pick a few ``seed" words or phrases (e.g. ``pet", ''dog'', ``cat" for the pet policy aspects) and expand the list by looking at those words' word2vec cosine distance neighborhoods. After our language specialists proofread the list, it is used to identify the relevant sentences via simple keyword matching. For the classification task we use a bag-of-ngrams linear model approach \cite{langford2007vowpal}.

As an example, Table \ref{tab:bsf:b} shows the BSF performance for ``parking availability" aspect in English $\rightarrow$ German translation. 

\begin{table}
\centering
\begin{subtable}{1\textwidth}
\centering
\begin{tabular}{r||c|c||c|c||c|c}
 & \multicolumn{2}{c}{\textbf{Precision}}  & \multicolumn{2}{c}{\textbf{Recall}}  & \multicolumn{2}{c}{\textbf{F1 Score}}\\
& EN & DE & EN & DE & EN & DE\\
\hline
Free parking & 0.97 & 0.96 & 0.93 & 0.94 & 0.95 & 0.95\\
Non-free parking\footref{nonfree} & 0.79 & 0.83 & 0.89 & 0.85 & 0.84 & 0.84\\
Not about parking& 1.00 & 0.99 & 1.00 & 1.00 & 1.00 & 0.99\\
\hline
Average & 0.97 & 0.96 & 0.97 & 0.96 & 0.97 & 0.96\\

\end{tabular}
\vspace{.1in}
\caption{Performance of English and German components of our BSF framework measured with a hold-out set of 500 examples.}\label{tab:bsf:a}
\end{subtable}

\quad

\vspace{.2in}
\begin{subtable}{1\textwidth}
\centering

\begin{tabular}{rc||ccc}
 & & \multicolumn{3}{c}{\textbf{German prediction}} \\
  &  &   Free parking &  Non-free parking\footref{nonfree}    & Not about parking   \\
\hline
\multirow{3}{*}{\rotatebox{90}{\textbf{\ \ English}}\ \rotatebox{90}{\textbf{prediction}}  } &   Free parking &   99.4\% &    0.5\%  & 0.1\%  \\
 &  Non-free parking\footref{nonfree} & 5.1\%    & 94.6\% &  0.3\%   \\
  & Not about parking  & $<0.1$\%     & $<0.1$\%   & 99.9\%
\end{tabular}
\vspace{.1in}
\caption{The result of applying BSF to our English/German corpus, expressed in matches normalized by the total English volumes. For example out of all English sentences which BSF annotated as ``Free parking", 99.4\% also get predicted as ``Free parking" in German, while 0.5\% of those get identified as ``Non-free parking"\footnote{\label{nonfree} Non-free parking can either be a sentence about clearly paid parking, or it can be something ambiguous as ``There is parking available nearby"} and 0.1\% as not about parking at all.}
\end{subtable}

\caption{Business-sensitive translation errors analysis for English-German pair for the ``parking availability" aspect.}\label{tab:bsf:b}
\end{table}

\subsection{BLEU score vs human-based metrics}\label{sect:q_bleu}

While BLEU score is very convenient to use because it can be computed automatically, the main metrics we really trust are human-based (see Section \ref{sect:q_human}). Here we look at how the BLEU scores from our English-German corpus size experiment of Section \ref{sect:opt_learn} are correlated with adequacy/fluency metrics.

\begin{figure}[h]
\includegraphics[scale=0.45]{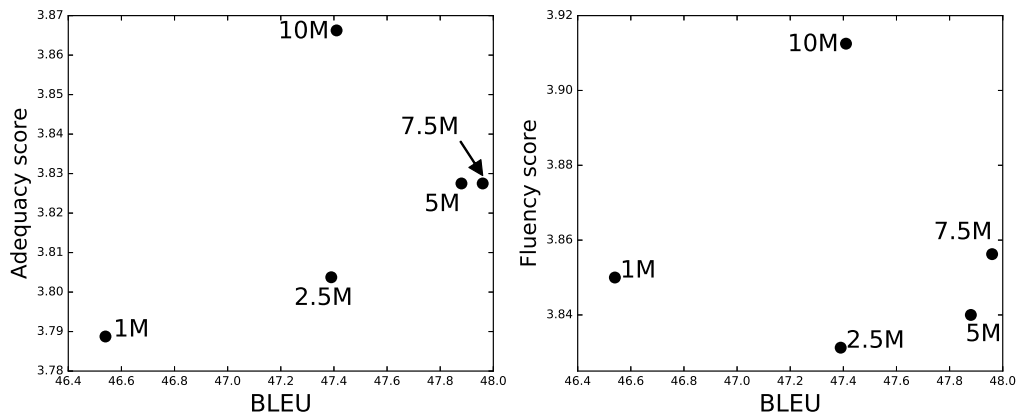}
\caption{BLEU against adequacy/fluency scores for English-German corpus size experiment from Section \ref{sect:q_human}}\label{fig:bleu_af}
\end{figure}

The results are shown in Figure \ref{fig:bleu_af}. The training with the corpus size of 10M clearly gives the best performance according to human evaluation, however this is not reflected in the BLEU score. As we can see the correlation between human metrics and BLEU score is rather tenuous. In particular, had we only looked at BLEU, we could have easily made the wrong conclusion about our experiment from Section \ref{sect:opt_learn}.

\section{Conclusion}
We have presented our approach to developing a large scale NMT system, specifically focusing on practical considerations. We presented the performance of different optimization strategies for model training in single- and multi-GPU environments. We found that a combination of Adam and SGD with learning rate decay works the best on a single GPU, and asynchronous SGD parallelization is a great strategy to dramatically speed up the training. We presented the advantages of BPE tokenization for machine translation and argued in favor of preprocessing named entities for better quality translation. Finally, we presented our approach of dealing with critical translation mistakes through our business sensitivity framework and argue that despite being the main metric in research, BLEU score alone can be a poor way of tracing MT system improvement.

In the future we are going to continue running optimization related experiments, particularly around better strategies for taking advantage of multiple GPUs. In order to leverage our massive monolingual corpora that are not translated, we are also focusing more on the research topics of model pre-training and similar techniques. Other important research topics to us are domain adaptation and user-generated content.

\section*{Acknowledgements}
We would like to thank our Language Specialists for providing invaluable human feedback and to Darina Kozlova for her important advice on human evaluation and for patiently coordinating all that work.

\newpage
\small

\bibliographystyle{apalike}
\bibliography{amta_paper}

\end{document}